\documentclass[10pt,twocolumn,letterpaper]{article}

\usepackage{iccv}
\usepackage{times}
\usepackage{epsfig}
\usepackage{graphicx}
\usepackage{amsmath}
\usepackage{amssymb}

\usepackage{graphicx}
\usepackage{amsmath}
\usepackage{amssymb}
\usepackage{booktabs}
\usepackage{multirow}
\usepackage{enumitem}
\usepackage{xcolor}
\usepackage{soul}
\usepackage{colortbl}
\usepackage[accsupp]{axessibility}

\definecolor{Gray}{gray}{0.9}

\usepackage[pagebackref=true,breaklinks=true,letterpaper=true,colorlinks,bookmarks=false]{hyperref}

\usepackage[capitalize]{cleveref}
\crefname{section}{Sec.}{Secs.}
\Crefname{section}{Section}{Sections}
\Crefname{table}{Table}{Tables}
\crefname{table}{Tab.}{Tabs.}

\iccvfinalcopy 

\def\iccvPaperID{6033} 

\ificcvfinal\fi

\newcolumntype{a}{>{\columncolor{Gray}}c}

\newcommand{\navcaptask}{Embodied Captioning}
\newcommand{\dataset}{ET-Cap}
\newcommand{\modelname}{CaBOT}

\begin{document}

\title{Explore and Tell: Embodied Visual Captioning in 3D Environments}

\author{Anwen Hu\textsuperscript{\rm 1}, Shizhe Chen\textsuperscript{\rm 2}, Liang Zhang\textsuperscript{\rm 1}, Qin Jin\textsuperscript{\rm 1}\thanks{Corresponding Author.}\\
  \textsuperscript{\rm 1}School of Information, Renmin University of China \\
  \textsuperscript{\rm 2}INRIA \\
  {\tt\small \{anwenhu,zhangliang00,qjin\}@ruc.edu.cn} \\
  {\tt\small shizhe.chen@inria.fr}}

\maketitle
\ificcvfinal\thispagestyle{empty}\fi

\begin{abstract}
While current visual captioning models have achieved impressive performance, they often assume that the image is well-captured and provides a complete view of the scene. In real-world scenarios, however, a single image may not offer a good viewpoint, hindering fine-grained scene understanding. To overcome this limitation, we propose a novel task called Embodied Captioning, which equips visual captioning models with navigation capabilities, enabling them to actively explore the scene and reduce visual ambiguity from suboptimal viewpoints. Specifically, starting at a random viewpoint, an agent must navigate the environment to gather information from different viewpoints and generate a comprehensive paragraph describing all objects in the scene. To support this task, we build the ET-Cap dataset with Kubric simulator, consisting of 10K 3D scenes with cluttered objects and three annotated paragraphs per scene. 
We propose a Cascade Embodied Captioning model (CaBOT), which comprises of a navigator and a captioner, to tackle this task. The navigator predicts which actions to take in the environment, while the captioner generates a paragraph description based on the whole navigation trajectory. 
Extensive experiments demonstrate that our model outperforms other carefully designed baselines. Our dataset, codes and models are available at \url{https://aim3-ruc.github.io/ExploreAndTell}.
\end{abstract}

\section{Introduction}
\label{sec:intro}

\begin{figure}
    \centering
    \includegraphics[width=1.0\linewidth]{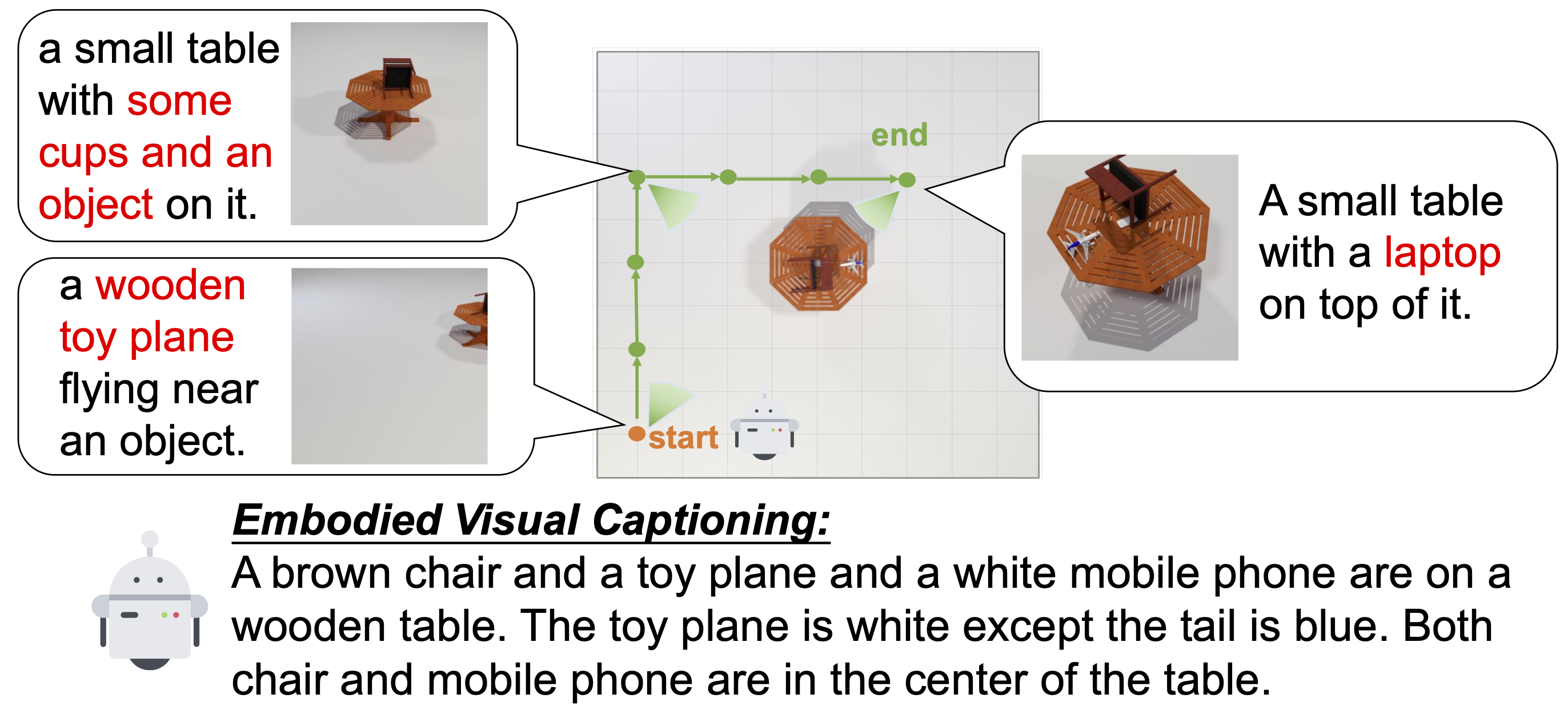}
    \caption{
    Existing visual captioning models generate a sentence to describe a single image, which often fail when the image is not well-captured at good viewpoints (the captions next to the images are generated by the state-of-the-art BLIP model~\cite{Li2022blip}).
    To generate more accurate and comprehensive visual descriptions, we propose a new Embodied Captioning task which allows agents to navigate the environment to reduce visual ambiguity of the scene.
    }
    \label{fig:intro_case}
\end{figure}

Visual captioning \cite{Chen2015cococap,Krause2017paragraphcap,Sidorov2020textcap,Gurari2020vizwizcap,Biten2019goodnews} is an essential vision-and-language task which aims to generate natural language descriptions of visual contents.
In recent years, many captioning models have been developed and achieved significant improvements in describing major objects and relationships in an image ~\cite{Xu2015sat,Anderson2018butd,pan2020x,Li2020oscar,Zhang2021vinvl,Li2022blip,Wang2022simvlm}.
However, the current models typically rely on well-captured images that provide a good viewpoint of the scene.
Unfortunately, in real-world scenarios, capturing such images may not always be feasible.
As illustrated in~\cref{fig:intro_case}, the initial position to capture an image may not provide a complete view of the scene, and a single image may not be sufficient to capture all objects within it,  potentially leading to incomplete or inaccurate visual captions.
To address this limitation, it is essential for visual caption models to navigate the environment actively and gather information from multiple viewpoints in order to generate more comprehensive and accurate visual captions.

In this paper, we propose a novel task called Embodied Captioning, which integrates such navigation ability into visual captioning. 
The task requires an agent, which starts at a random viewpoint in a 3D environment, to navigate the environment to reduce visual ambiguity of the scene, and finally generate a comprehensive paragraph that describes all objects in the scene.
Different from existing navigation tasks~\cite{Anderson2018vln,Das2018eqa,Yu2019mteqa,Qi2020reverie,Li2022revece}, our Embodied Captioning task does not explicitly define a target location to navigate. Instead, we have an implicit target which is to accurately recognize all objects along with their attributes and relationships in the scene as soon as possible.

To support the Embodied Captioning task, we build a high-quality 3D dataset \dataset~with manually annotated paragraph descriptions. 
Leveraging high-quality 3D object assets from ShapeNet~\cite{Chang2015shapenet} and GSO~\cite{Downs2022gso}, we use Kubric~\cite{Greff2022kubric} to construct 10,000 scenes. 
For each scene, three annotators are provided with 20 images from different viewpoints of the scene and asked to write a detailed paragraph to describe all visible instances. 
We also require the annotators to select images with good viewpoints among all the image candidates.
Although our dataset is based on synthetic scenes with limited obstacles, it still presents significant challenges for Embodied Captioning. The agent only receives an RGB image of a restricted field of view at each step without any localization information, such as the location of the agent and objects.
Therefore, a model must be equipped with long-term memories of previous visual observations and actions in order to efficiently explore the environment, accurately recognize objects, and generate fine-grained scene descriptions.

To address these challenges, we propose a Cascade Embodied Captioning model (CaBOT), which consists of a History-aware Navigator and a Trajectory-aware Captioner. 
The navigator leverages histories of both observed images and performed actions to predict the next action via a transformer model. 
The captioner is fed with all images in the predicted trajectory and utilizes a bi-level cross attention over spatial and temporal dimensions of the image sequence to generate a paragraph. 
The proposed CaBOT model outperforms carefully designed baselines for the challenging Embodied Captioning task.
Nevertheless, there is still much room to improve compared to human performance, such as joint modeling the navigation and captioning process.

In summary, our contributions are three-fold:
\parskip=0.1em
\begin{itemize}[itemsep=0.1em,parsep=0em,topsep=0em,partopsep=0em]
    \item We propose a novel and challenging  \navcaptask~task which requires agents to explore in 3D environments to generate better visual descriptions. 
    \item A high-quality dataset is constructed to benchmark the \navcaptask~task, with 10K synthetic 3D scenes and 24K manually annotated good viewpoints and 30K paragraph descriptions.
    \item We present a Cascade Embodied Captioning model which incorporates  navigation histories for captioning, providing a strong starting point for future work.
\end{itemize}
\section{Related Work}
\label{sec:rela}

\noindent\textbf{Visual Captioning.} Visual captioning aims to describe visual contents with natural language. A variety of visual captioning tasks have been proposed such as image captioning \cite{Chen2015cococap,Krause2017paragraphcap,Sidorov2020textcap,Gurari2020vizwizcap,Biten2019goodnews}, video captioning \cite{Xu2016msrvtt, Xu2016msrvtt, Zhou2018youcook, Wang2019vatex,wang2019youmakeup} and 3D captioning \cite{Chen2021scan2cap}.
In image captioning, a model should describe major objects, attributes, relations \cite{Chen2015cococap, young2014image, Krause2017paragraphcap, agrawal2019nocaps, Chen2020graphcontrol} or even named entities \cite{Biten2019goodnews} and scene texts \cite{Sidorov2020textcap, Hu2021qctextcap} in the image. 
Most of the image captioning datasets \cite{lin2014microsoft,Krause2017paragraphcap,Sidorov2020textcap,Biten2019goodnews} use high-quality web images. Such dataset bias results in poor generalization when an image is not well taken as shown in VizWiz-Caption dataset \cite{Gurari2020vizwizcap} which contains photos taken by the blind.
Video captioning tasks \cite{Xu2016msrvtt, Zhou2018youcook, Wang2019vatex,wang2019youmakeup} take a video as input, and focus more on the actions or events. 3D captioning task \cite{Chen2021scan2cap} is designed to describe a 3D indoor scene with almost complete point clouds as input. 
Much progress has been made to improve the captioning performance such as attention mechanism \cite{Xu2015sat,Anderson2018butd,Hu2020icecap}, transformer architecture\cite{pan2020x,Yuan2022xtrans2cap,song2021onestage} and pretraining framework \cite{DBLP:conf/aaai/ZhouPZHCG20,Li2022blip,Li2020oscar,Yang2021tap,Wang2022simvlm}.
In this work, we aim to push the standard visual image captioning tasks one step further, which do not passively receive an visual input, but should actively explore in the environment to obtain better visual observations to describe.

\noindent\textbf{Embodied Vision and Language Tasks.} 
Growing research attentions have been paid to embodied vision and language tasks \cite{Anderson2018vln,Das2018eqa,Yu2019mteqa,Qi2020reverie,Tan2020esd,Li2022revece} in recent years, which require agents to perform actions to achieve various goals specified in natural language. 
Vision-and-Language Navigation \cite{Anderson2018vln} requires an agent to follow natural language instructions to navigate to a target place. Embodied visual referring expression \cite{Qi2020reverie,Li2022revece} provides object-oriented high-level instructions and requires both navigation and object grounding. In embodied question answering \cite{Das2018eqa,Yu2019mteqa}, a question is provided to the agent such as `What room is the microwave located in?', and the agent should firstly find relevant visual information and then generate the answer. 
Compared with the above tasks, our proposed Embodied Captioning does not provide explicit goals for navigation.
In contrast, there is only an implicit goal that the agent should generate more informative captions after its navigation.

The most similar work to ours is embodied scene description  (ESD) \cite{bigazzi2021explore, bigazzi2023embodied,Tan2020esd}. 
Our work is distinguished from ESD in two main aspects.
Firstly, the main goal of navigation in ESD is to enlarge the explored area or to find a single viewpoint that captures more objects, while our navigation goal is more entangled with visual captioning to enable to generate accurate and comprehensive captions as soon as possible.
Secondly, existing ESD work simply uses pretrained image captioning models and does not have an in-domain  dataset to benchmark the task with systematic evaluation.
In this work, we construct a high-quality dataset with human-annotated viewpoints and paragraph descriptions and we also provide multiple strong baselines.

\section{\dataset~Dataset}

We build a high-quality synthetic dataset ET-Cap with human annotations to teach agents how to \textbf{E}xplore and \textbf{T}ell in 3D environments.

\begin{figure*}
    \centering
    \includegraphics[width=1.0\linewidth]{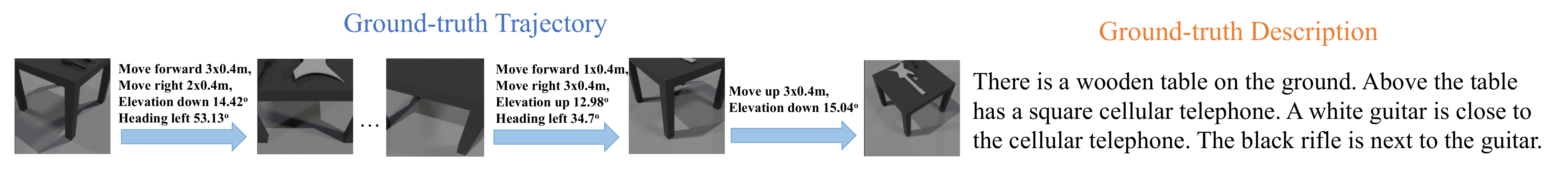}
    \caption{An example of ground-truth trajectory and corresponding description in \dataset.}
    \label{fig:train_sample}
\end{figure*}

\subsection{3D Scenes Simulation}
\label{sec:3d_sim}
3D scenes in ET-Cap are constructed with Kubric \cite{Greff2022kubric}, a recently proposed framework that enables the generation of photo-realistic scenes. 
We use Kubric to place multiple objects in a scene and render images. 3D object assets come from ShapeNet \cite{Chang2015shapenet} and Google Scanned Objects \cite{Downs2022gso}.

Each scene in \dataset~is constructed following three steps: \emph{Instance Resizing}, \emph{Instance Selection}, and \emph{Instance Placing}. Specifically, in order to make the scene more realistic, we first resize 3D object assets to their common sizes in real life. 
Since the indoor environment is often organized by arranging small objects around big furniture, we then randomly choose a big-size furniture instance (called a base instance) and multiple relatively smaller instances (called placing instances). 
Finally, to ensure the diversity of instance arrangement, we place instances by performing a physic simulation. The base instance is first placed at the center of the scene. The other instances are then dropped above the base instance, leading to diverse poses of instances across scenes because of collision.
More details can be found in the supplementary material.

\noindent\textbf{Navigation Space.} 
We discretize the whole environment as a 3D grid world with the total size of $8\times8\times4~m^3$  (length$\times$width$\times$height) and each grid size of $0.4 \times 0.4 \times 0.4~m^3$.
There are on average 4,838 navigable grids in a scene removing the grids occupied by instances.
Agents in the 3D grid world can perform five types of actions relative to their current pose: \verb|forward-backward move| (stop, forward, backward), \verb|left-right move| (stop, left, right), \verb|up-down move| (stop, up, down), \verb|heading rotate| (stop, left, right) and \verb|elevation rotate| (stop, up, down). The maximum distance traveled in each direction is 1.6m. Each action type involves a combination of a direction classification and a magnitude regression task, \eg, move forward for 0.8$m$,  rotate horizontally to the left for 30 degrees.

\subsection{Manual Annotation}
After the automatic scene construction, we recruit expert annotators from Appen Platform\footnote{\url{https://www.appen.com/}} to select good viewpoints and write detailed descriptions for each scene.

\noindent\textbf{Viewpoint Selection.} 
Explicitly annotating trajectories for Embodied Captioning is time-consuming, subjective, and less flexible due to dependency on initial positions.
Therefore, we ask annotators to select good viewpoints for each scene at which the scene can be well-captured. This allows us to automatically generate trajectories from random initial positions to those good viewpoints (Section~\ref{sec:traj_generation}).
To be specific, as good viewpoints are always a certain distance away from instances in both horizontal and vertical directions, we first sample 20 candidate viewpoints in a highly probable spatial range of the grid world and then set the camera to look at the center of the scene to render images.
To help annotators understand the scene, we provide category labels of instances in the scene.
The annotators can select \emph{multiple} good viewpoints as long as the image provides a clear view of (almost) all objects in the scene.

\noindent\textbf{Paragraph Annotation.} 
We ask annotators to write descriptions about objects in the scene, mentioning categories, attributes, and spatial relationships of instances. 
To ensure that the description is comprehensive, annotators are required to write a paragraph in which the number of sentences roughly matches the number of instances. 

For each scene, there are three independent expert annotators recruited to select viewpoints and write paragraphs. Besides, an extra expert inspector is arranged to check the annotation quality. More details about the annotation platform, annotation cost, and annotation samples can be found in the supplementary material.

\subsection{Trajectory Generation}
\label{sec:traj_generation}

We generate ground-truth trajectories based on the annotated good viewpoints and ground-truth 3D environment information in each scene.
Specifically, we take a good viewpoint as the target position and randomly select three start positions. 
In contrast to previous navigation works~\cite{Anderson2018vln,Das2018eqa,Tan2020esd} which use a fixed move step size of 0.25m, we allow our agent to move more flexibly and cover longer distances, with a maximum length of 1.6m, to reduce trajectory length.
We employ the Dijkstra algorithm to generate the shortest movement path in the 3D space from the start to the target position.
Then, we generate an optimal rotation action of the camera after each move step of the agent.
To ensure that most instances are visible, we directly set the camera to look at the center of the scene, as we can use the perfect agent's location and object locations for ground-truth trajectory generation. According to the camera view, position changes are converted to three types of move actions. 
We merge the move action and the camera rotation action at each step, resulting in five ground-truth actions at each step: 3 movements with 3 step lengths, a camera heading rotation with direction and angle, and a camera elevation rotation with direction and angle.
\cref{fig:train_sample} shows an example of a ground-truth trajectory and its paragraph description.

\subsection{Dataset Statistics}

\begin{figure}
    \centering
    \includegraphics[width=0.9\linewidth]{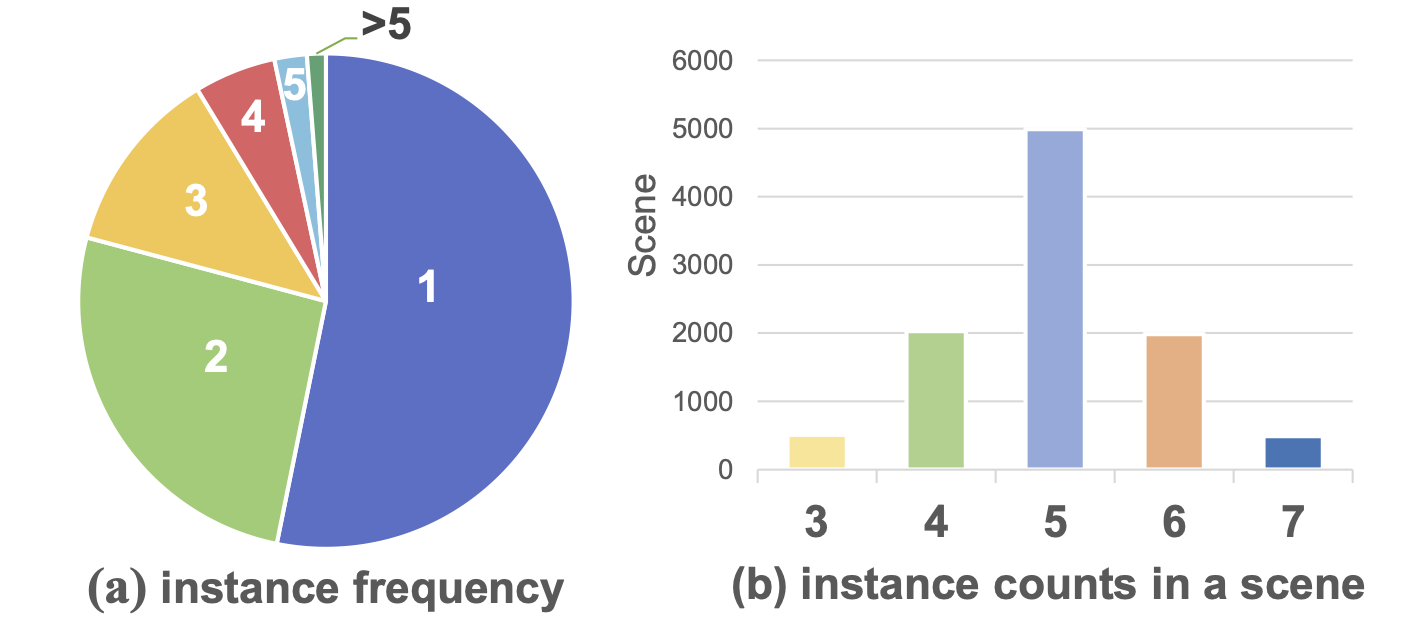}
    \caption{Instance frequency distribution (a) and distribution of the number of instances in a scene (b).}
    \label{fig:freq_and_instance_num}
\end{figure}

\noindent\textbf{Scene Statistics.} 
Our ET-Cap dataset consists of 10,000 scenes and 27,520 unique instances, which include 7,517 base instances and 20,003 placing instances across 10 and 47 object categories respectively. 
The detailed category distribution is provided in the supplementary material.
Notably, about 53\% of the instances appear only once in the dataset, as shown in~\cref{fig:freq_and_instance_num}(a), which can prevent models from overfitting the instances. 
As shown in~\cref{fig:freq_and_instance_num}(b), the number of instances in a scene varies from 3 to 7.

\noindent\textbf{Trajectory Statistics.} 
We generate 72,594 trajectories with an average of 6.13 viewpoints and a mean length of 6.3m. 
The ground-truth trajectories exhibit no significant bias in the action direction. The `move down' direction is relatively less frequent than the `move up' direction because most good viewpoints are in higher positions than the height of the objects, which is consistent with our intuition.

\begin{table}
\caption{Statistics of captions in our \dataset~ compared to other visual captioning datasets.}
    \label{tab:cap_datasets}
    \footnotesize
    \tabcolsep=0.06cm
    \centering
    \begin{tabular}{cccccc}\toprule
    Statistics & COCO~\cite{Chen2015cococap} & ImPG~\cite{Krause2017paragraphcap} & Nr3D~\cite{Achlioptas2020nr3d} & ScanR~\cite{Chen2021scan2cap} & \dataset \\
     \midrule
    \# of words & 11.3 & 68.5 & 12.4 & 20.2 & 56.3 \\
    \# of sentences & 1.8 & 6.7 & 1.7 & 3.0 & 5.9 \\
    adjectives (\%) & 6.54 & 9.84 & 8.73 & 9.38 & 17.01 \\
    nouns (\%) & 32.30 & 25.13 & 26.65 & 24.12 & 24.62 \\
    spatial words (\%) & 9.72  & 8.51 & 11.87 & 10.15 & 10.84 \\
    \bottomrule
    \end{tabular}
\end{table}

\noindent\textbf{Caption Statistics.} We collect 30,000 captions in total. A caption contains an average of 5.9 sentences and 50.2 words. The average number of adjectives, nouns and spatial relation words (e.g. `left', `right', `center') in a caption are 9.6, 13.8, and 6.1, respectively. As shown in \cref{tab:cap_datasets}, our captions are more descriptive than those in existing visual captioning datasets, with a much higher ratio of adjectives.
They also contain slightly more spatial words compared to traditional image captions~\cite{Chen2015cococap,Krause2017paragraphcap} and competitive to 3D object referring expression datasets~\cite{Achlioptas2020nr3d,Chen2021scan2cap}.

\begin{table}
\caption{Statistics of \dataset~training, validation and test sets.}
    \label{tab:split}
    \footnotesize
    \tabcolsep=0.2cm
    \centering
    \begin{tabular}{ccccc}
    \toprule
    \textbf{Split} & \textbf{Subset} & \textbf{Scene} & \textbf{Caption} & \textbf{Trajectory} \\
    \midrule
    Training & - & 8,324 & 24,972 & 60,474 \\
    \midrule
    \multirow{4}*{Validation} & Common & 300 & 900 & 2,178 \\
    ~ & Novel Instance & 300 & 900 & 2,176 \\
    ~ & Novel Category & 238 & 714 & 1,638 \\
    ~ & Total & 838 & 2,514 & 6,012 \\
    \midrule
    \multirow{4}*{Test} & Common & 300 & 900 & 2,178 \\
    ~ & Novel Instance & 300 & 900 & 2,223 \\
    ~ & Novel Category & 238 & 714 & 1,707 \\
    ~ & Total & 838 & 2,514 & 6,108 \\
    \bottomrule
    \end{tabular}
\end{table}

\noindent\textbf{Dataset Splits.}  
We split our \dataset~into training, validation, and test sets based on scenes. \cref{tab:split} shows the statistics of each set. 
To test the generalization ability of agents, we further divide the validation and test sets into three subsets: a) \emph{Common subset}, in which instances in the scene have all been seen in the training set; b) \emph{Novel Instance subset}, in which the categories are all covered in the training set, but more than 50\% of the instances in the scene do not appear in the training set; c) \emph{Novel Category subset}, in which each scene contains at least 1 novel object category that has not been seen in the training set.

\label{sec:dataset}
\section{Cascade Embodied Captioning Model}
\begin{figure*}
    \centering
    \includegraphics[width=1.0\linewidth]{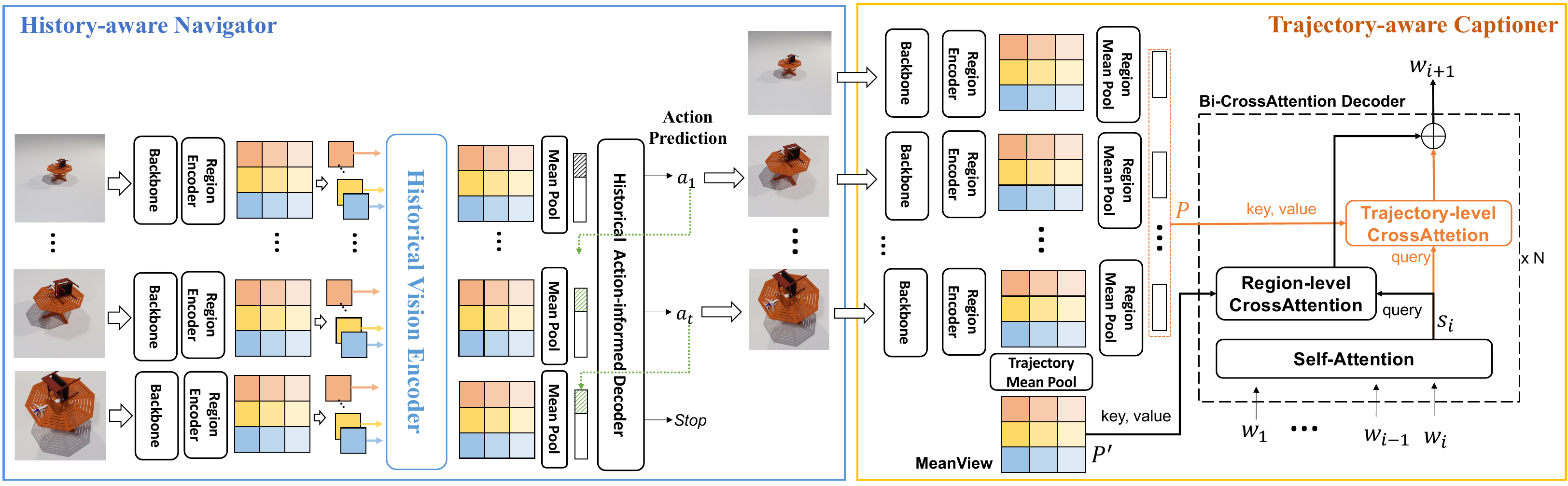}
    \caption{The overall architecture of \textbf{Ca}scade em\textbf{BO}died cap\textbf{T}ioning (CaBOT) model. It consists of a History-aware Navigator  and a Trajectory-aware Captioner. The navigator navigates from a start viewpoint in a 3D environment. The captioner generates a paragraph to describe the scene according to the trajectory of navigator. }
    \label{fig:model}
\end{figure*}

In the \navcaptask~task, the agent is initialized at a bad viewpoint and can only receive an RGB image at each step. It should navigate in the 3D environment to comprehend the scene and generate a paragraph describing all objects at the end of the navigation.
We propose a \textbf{Ca}scade em\textbf{BO}died cap\textbf{T}ioning (\modelname) model to tackle the task. As shown in~\cref{fig:model}, \modelname~is composed of a History-aware Navigator and a Trajectory-aware Captioner. The navigator utilizes an episodic history to explore the environment, while the captioner generates paragraphs using the whole trajectory produced by the navigator. 

\subsection{History-aware Navigator}
Given the observed RGB image at each step, the navigator must generate 5 types of actions: forward-backward movement, left-right movement, up-down movement, elevation rotation, and heading rotation.
For each action type, it needs to predict both the direction and magnitude.
Once the predictions are made, the agent will move to a new position and adjust its camera orientation accordingly. It stops until all the values of the types of actions are `stop'.

Our navigator consists of a backbone to extract image features where we use a ResNet~\cite{He2016resnet}, a region encoder to enhance spatial modeling within each image, a historical vision encoder to enhance temporal modeling across images, and a historical camera-informed decoder to predict actions.

\noindent\textbf{Region Encoder.} During navigation, to know which direction is possible to capture more visual information of instances, it is necessary to compare the region difference of the current image. Therefore, we apply a Transformer layer \cite{vaswani2017attention} with self-attention over the backbone for region-level encoding. For the image $v_{t}$ rendered from the $t^{th}$ viewpoint, its region-level features are denoted as  $H_{t}=[h^1_{t},h^2_{t}, ..., h^R_{t}]$, where $R$ is the region sequence length.
 
 \noindent\textbf{Historical Vision Encoder.} Besides spatial relation within the current image, historical vision information also helps to decide the next actions. For example, observing the change of the central region from blank to one containing partial instances helps to confirm continued movement in the same direction. Thus, we further apply a Historical Vision Encoder to learn the temporal relation of the same region across images at different steps as follows:
 \begin{gather}
    \tilde{h^{r}_{t}}={\rm Trans_{<}}([h^{r}_{1},.., h^{r}_{t-1},h^{r}_{t}]), 1 \leq r \leq R, \label{eq:vm_en}
\end{gather}
 where ${\rm Trans_{<}(
 \cdot)}$ is a Transformer layer with causal-masked self-attention mechanism, which means only previous image information can be attended. 
 
\noindent\textbf{Historical Action-informed Decoder.}
In addition to historical vision information, historical action information (direction and magnitude) is also helpful for deciding the next actions because it records the viewpoint change during previous navigation. Therefore, we design a Historical Action-informed Decoder leveraging both vision and action information to predict the next actions as follows:
  \begin{align}
    & c_t = W^{c}a_{t-1}+b^{c}, \\
    & o_{t} = [\begin{matrix}\frac{1}{R}\sum_{r=1}^{R} \tilde{h^r_t} \end{matrix};c_t], \\
    & \bar{o_t}={\rm Trans_{<}}([o_1,.., o_{t-1},o_t]),\label{eq:nav_de}
\end{align}
where $a_{t-1}$ is a 10D vector consisting of the five independent action classes and magnitudes in the previous step. $W^{c}$ and $b^{c}$ are trainable parameters, $[;]$ means concatenation.

With $\bar{o}_t$ as the input, we apply linear layers to predict the next action $a_t$ including the direction and magnitude for each of the five types of actions.
We use a softmax function for the direction classification and a sigmoid function for magnitude regression at the end of the linear layer.

\subsection{Trajectory-aware Captioner}
The Trajectory-aware Captioner takes images $V=[v_1,v_1,...,v_T]$ observed in the navigation trajectory as input, and outputs a paragraph to describe the scene. 
Due to the importance of accurate object recognition for visual captioning~\cite{Anderson2018butd,Li2020oscar}, we train a DETR~\cite{Carion2020detr} model to detect objects (more details can be found in our supplementary material) and use its backbone ResNet~\cite{He2016resnet} to initialize the backbone in our captioner. 
Similar to the navigator, the captioner also applies a Region Encoder which converts the image sequence into $\mathcal{H}=[H_{1}, ...,H_{T}]$, where $H_t=[h^1_{t},h^2_{t}, ..., h^R_{t}], 1 \leq t \leq T$, $T$ is the trajectory length, and $R$ is the region sequence length.

\noindent\textbf{Bi-CrossAttention Decoder}
In order to gather visual information from different steps in the navigation trajectory, we propose a Bi-CrossAttention Decoder to leverage all images at both the region level and trajectory level for captioning.

We first perform a region-level mean pooling and a trajectory-level mean pooling on $\mathcal{H}$ as follows:
 \begin{gather}
    p_{t} = \begin{matrix}\frac{1}{R}\sum_{r=1}^{R} h^r_t \end{matrix}, 1 \leq t \leq T, \\
    p'_{r} = \begin{matrix}\frac{1}{T}\sum_{t=1}^{T} h^r_t \end{matrix}, 1 \leq r \leq R,
\end{gather}
$P=[p_1,..,p_T]$ and $P'=[p'_{1},..,p'_{R}]$ are pooled trajectory-level and region-level visual features, respectively. 

At the $i^{th}$ decoding step of paragraph generation, the decoder first performs causal-masked self-attention on previous words to keep text coherence. Then it applies two cross-attention layers to select relevant visual information at the region level and trajectory level as follows:
 \begin{align}
   &s_{i} = {\rm Trans_{<}}([w_{1},w_{2}, ...,w_{i}]),\\
   &\alpha = {\rm Softmax}([s_{i}^\intercal p'_1,..., s_{i}^\intercal p'_R]), \\
   &\beta = {\rm Softmax}([s_{i}^\intercal p_1,..., s_{i}^\intercal p_T]), \\
   & y_{i} = \begin{matrix} \sum_{r=1}^R \alpha_{r} p'_{r} \end{matrix} + \begin{matrix} \sum_{t=1}^T \beta_{t}p_{t} \end{matrix},
\end{align}
where $s_{i}$ is the output of casual-masked self-attention for word $w_{i}$. Projection matrices in attention layers are omitted for equation simplicity.
$y_{i}$ is finally added with $s_{i}$ and used to predict the next word $w_{i+1}$ with a classification layer.

\subsection{Training and Inference}
We train the History-aware Navigator and the Trajectory-aware Captioner separately.
Both modules leverage images in ground-truth trajectories as the input for imitation / supervised learning. The loss function of the navigator is the summation of each type of action. 
Assume $\mathcal{L}^{cls}$ is the cross-entropy loss for action direction classification and $\mathcal{L}^{reg}$ is the L1 loss for action magnitude regression,
the navigation loss is: $\mathcal{L}=\mathcal{L}^{cls}+2.0\mathcal{L}^{reg}$.
The captioner (including its backbone) is optimized by the cross-entropy loss of word prediction~\cite{Anderson2018butd,Li2020oscar, Xu2015sat}.
More implementation details can be found in the supplementary material.

At the inference phase, \modelname~performs Embodied Captioning in a cascade manner. Given a starting viewpoint, the navigator first generates a trajectory. The captioner then leverages images from the predicted trajectory to generate a paragraph description.

\section{Experiments}
\label{sec:experiments}

\begin{table*}
\caption{Captioning performance with oracle trajectories.`
`single' means only utilizing a single view at the end of the trajectory for region-level cross-attention, and `multi' means using the average of multi-view images in the whole trajectory.
`Human' denotes using one human-written caption as the candidate and the other two human-written captions as references. }
    \label{tab:oracle_cap_experiments}
    \footnotesize
    \tabcolsep=0.2cm
    \centering
    \begin{tabular}{c|ccc|ccccccaa}
    \toprule
    ~  & Region-level & Trajectory-Level & DETR Init &\multirow{2}*{BLEU1} & \multirow{2}*{BLEU2} & \multirow{2}*{BLEU3} & \multirow{2}*{BLEU4} & \multirow{2}*{METEOR} & \multirow{2}*{ROUGE-L} & & \\
     ~  & CrossAtt & CrossAtt & Backbone & ~ & ~ & ~ & ~ & ~ & ~ &  \multirow{-2}*{CIDEr}  & \multirow{-2}*{SPICE} \\
    \midrule
    r1  & \multicolumn{3}{c|}{Template-based}  & 56.79 & 38.32 & 23.55 & 13.22 & 21.10 & 38.44 & 26.41 & 18.25 \\
    \midrule
    r2 & single &$\times$ & $\times$ & 62.88 & 47.83 & 35.54 & 25.46 & 20.76 & 44.99 & 29.34 & 14.22 \\
    r3 & multi & $\times$ & $\times$  & 62.35 & 46.76 & 34.15 & 24.74 & 22.33 & 43.34 & 28.35 & 16.25 \\
    r4  & $\times$ & $\checkmark$ & $\times$  & 66.36 & 50.49 & 36.89 & 26.24 & 23.81 & 44.59 & 39.49 & 18.28 \\
    r5 & multi & $\checkmark$ & $\times$ &66.26 & 51.00 & 37.86 & 27.58 & 24.21 & 45.72 & 42.10 & 19.22 \\
    r6 & multi & $\checkmark$ & $\checkmark$  &\textbf{66.79} & \textbf{51.44} & \textbf{38.02} &  \textbf{27.62} & \textbf{24.50} & \textbf{46.00} & \textbf{45.24} & \textbf{20.45} \\
    \midrule
    r7 & \multicolumn{3}{c|}{Human}   & 73.47 & 55.73 & 41.20 &  29.84 & 29.84 & 49.24 & 96.72 & 39.73 \\
    \bottomrule
    
    \end{tabular}
\end{table*}

\begin{table*}
\caption{Embodied Captioning performance using different navigators and our proposed captioner (r6 in Tab.~\ref{tab:oracle_cap_experiments}).
}
    \label{tab:navi_experiments}
    \footnotesize
    \centering
    \begin{tabular}{l|cc|cccccc|cccc}
    \toprule
    \multirow{2}*{\textbf{}} & Action & Historical &  \multicolumn{6}{c|}{\textbf{Navigation Evaluation}} & \multicolumn{4}{c}{\textbf{Captioning Evaluation}} \\
     ~ & history & Vision Encoder & NE$\downarrow$ & IS$\uparrow$ & SS$\uparrow$ & NE$^{l}$$\downarrow$ & IS$^{l}$$\uparrow$ & SS$^{l}$$\uparrow$ & CIDEr$\uparrow$ & SPICE$\uparrow$ & CIDEr$^{l}$$\uparrow$ & SPICE$^{l}$$\uparrow$ \\
    \midrule
    r1  &\multicolumn{2}{c|}{Rule-based} & 4.42 & 45.25 & 30.31 & 11.24 & 17.79 & 11.53 & 33.90 & 16.54 & 15.80 & 7.69   \\
    \midrule
    r2 & $\times$ & $\times$   & 3.81 & 69.85 & 69.62 & 5.23 & 56.02 & 55.99 & 40.59 & 19.35 & 33.35 & 15.96 \\
    r3 & $\times$ & $\checkmark$ & 3.81  & 69.76  & \textbf{69.92}  & 5.00 & 57.70  & 57.75 & \textbf{40.91} & \textbf{19.45} & 34.44 & 16.45\\
    r4 & $\checkmark$ & $\checkmark$ & \textbf{3.78} & \textbf{70.15} & 69.58 & \textbf{4.89} & \textbf{58.54} &  \textbf{58.13} & 40.83 & 19.43 & \textbf{34.88} & \textbf{16.62} \\ 

    \bottomrule
    
    \end{tabular}
\end{table*}

\subsection{Evaluation Metrics}

\noindent\textbf{Captioning evaluation.}
We employ common captioning metrics to evaluate the quality of generated paragraphs in our \navcaptask~task, including BLEU4~\cite{papineni2002bleu}, METEOR~\cite{banerjee2005meteor}, ROUGE-L~\cite{lin2004rouge}, CIDEr~\cite{vedantam2015cider} and SPICE~\cite{anderson2016spice}.
The CIDEr and SPICE scores are emphasized because they have been demonstrated to be aligned with human evaluations better for image captions. 
CIDEr puts more weight on rare words, while SPICE better measures the semantic similarity of objects, object-attribute pairs, and object-relation triplets.
We compare each generated paragraph against three human-written descriptions as references.
To promote efficient exploration of the scene, we factor in trajectory length when evaluating descriptions.
Specifically, we calculate a weight $r^l=L_{gt}/\text{max}(L_{gt}, L_{pred})$, where $L_{gt}$ and $L_{pred}$ are the lengths of the ground-truth and predicted trajectories, respectively.
We then multiply conventional captioning metrics with $r^l$ to obtain length-penalized metrics, denoted as $*^{l}$ for each captioning metric.

\noindent\textbf{Navigation evaluation.}
We propose three types of evaluation metrics to separately measure the navigation quality.
The first one \textbf{N}avigation \textbf{E}rror (NE) is commonly adopted in previous embodied navigation works~\cite{Anderson2018vln,Das2018eqa}. It measures the minimum Manhattan distance in meters between a predicted viewpoint and the manually selected good viewpoints in a scene. 
However, as the selected good viewpoints are only a subset of good viewpoints, NE is a pessimistic metric of the navigation quality.
Therefore, we introduce the other two metrics that directly evaluate the quality of the image rendered at a predicted viewpoint: \textbf{I}mage \textbf{S}imilarity (IS) and \textbf{S}egmentation Similarity (SS). 
IS measures the visual similarity using a CLIP \cite{Radford2021clip} vision encoder:
\begin{gather}
    {\rm IS}(v, \mathcal{V})=\mathop{\max}_{v' \in \mathcal{V}}\frac{\cos({\rm CLIP}(v), {\rm CLIP}(v'))-\gamma}{1-\gamma},\label{eq:is}
\end{gather}
where $\mathcal{V}$ are images in good viewpoints, $v$ refers to an image in a predicted viewpoint, ${\rm CLIP}(\cdot)$ refers to the CLIP \cite{Radford2021clip} vision encoder clip-ViT-B-32, $\gamma$ is set to 0.7 to reduce the influence of background.
The SS metric measures the similarity of semantic segmentation between $v$ and $\mathcal{V}$:
\begin{gather}
    {\rm SS}(v, \mathcal{V})=\mathop{\max}_{v' \in \mathcal{V}}\frac{1}{N_C}\sum_{c\in C}\min(\frac{A(v, c)}{A(v', c)}, 1.0),\label{eq:ss}
\end{gather}
where $C$ refers to the set of categories covered in $v'$, $N_C$ refers to the number of categories in $C$, $A(\cdot, c)$ computes the area ratio of category $c$ in the image. 

As images in the whole trajectory might provide useful visual information about the scene, we evaluate the quality of the whole trajectory rather than only the stop viewpoint or best viewpoint. 
Specifically, for all the navigation metrics, we evaluate the score at each step in the trajectory and take the average as the final score. 
Similar to caption evaluation, we introduce trajectory length into navigation evaluation.
We multiply NE by $1/r^{l}$ to obtain NE$^{l}$, and multiply IS and SS score by $r^{l}$ to obtain IS$^{l}$ and SS$^{l}$, respectively.

\subsection{Captioning Ablation with Oracle Trajectory}
\label{sec:results_ec_oracle_traj}

In this section, we evaluate the effectiveness of the proposed captioner given ground-truth trajectories.
The compared baselines include a template-based method and different variants of our proposed captioner.
In the template-based method, we carefully design paragraph templates and insert automatically detected objects from images in the trajectory into the template (more details can be found in the supplementary material). 
In the variants of our model, we ablate the contributions of region-level cross-attention, trajectory-level cross-attention, and backbone initialization.

\cref{tab:oracle_cap_experiments} presents the comparison results.
Firstly, all models outperform the template-based method, indicating the rule-based method is not competent enough for describing the 3D scene in detail. Secondly, as the end view in an oracle trajectory is the image in good viewpoints annotated by humans, r2 achieves comparable performance with r3. However, it's still worse than r4, showing the importance of merging visual information at the trajectory level for scene description.
Additionally, by combing both region-level and trajectory-level cross attention, the captioner achieves better performance (r5 vs r4), indicating region-level features and trajectory-level features are complementary for paragraph generation. Furthermore, by initializing the backbone with ResNet from the object detection model, the captioner performs best (r6), showing knowledge about object detection is also fundamental to describing 3D environments. Finally, there is a large gap (e.g., 51.48 on CIDEr) between model performance and human performance, indicating there is still a large room to improve on our dataset.

\subsection{Embodied Captioning Results}
\label{sec:results_ec_pred_traj}

\noindent\textbf{Main results.}
In this section, we combine the best captioning model (r6) in~\cref{tab:oracle_cap_experiments} with automatically predicted trajectories by a navigator.
We compare the proposed navigator with its variants and a carefully designed rule-based method.
The rule-based method primarily determines the direction of movement by analyzing the distribution of object areas within the current field of view. Specifically, to decide movement directions, we adopt a cropping strategy that splits the current image into upper and lower parts of equal size. If the upper/lower part contains more instances, the agent will move up/down accordingly. We also divide the image into equal-sized left, middle, and right parts. The agent will move forward/left/right depending on where most instance area is located.
Moreover, if the agent moves forward, we will examine whether this results in fewer instances being visible. If so, the agent will move backward. Once the agent reaches a new position, we make it turn around by four 90$^{o}$ turns to the left in order to determine the best orientation.
The rule-based approach stops when the maximum trajectory length is reached.

\cref{tab:navi_experiments} presents both the navigation performance and the captioning performance.
Firstly, rule-based navigation is shown to be inferior to variants of CaBOT navigator, which suggests that navigating for better scene descriptions generation in \dataset~is not a naive task. 
Secondly, with historical vision information during decoding, the navigator achieves better performance (r3 vs r2), which indicates that visual difference across time at the same image region is helpful for navigation.
Besides, introducing historical action encoding further improves navigation performance (r4 vs r3), suggesting that the previous action trajectory is also beneficial to predict the next action. 
Finally, better navigation leads to better scene description, verifying that navigation ability is crucial for effectively embodied captioning.

\begin{figure*}[h!]
    \centering
    \includegraphics[width=0.95\linewidth]{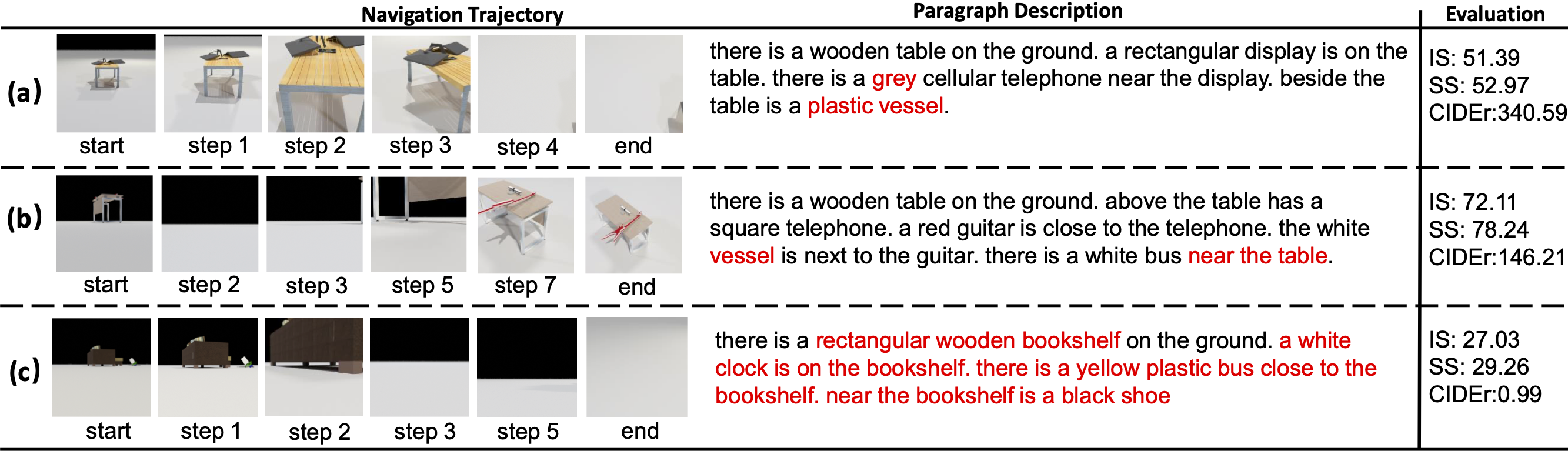}
    \caption{Qualitative results of \modelname~on Embodied Captioning test set. Wrong descriptions are marked in {\color{red}red}.}
    \label{fig:case_study}
\end{figure*}

\noindent\textbf{Generalization analysis.}
We evaluate the generalization ability of \modelname~on three subsets of the test set. As shown in \cref{tab:subset_experiments}, \modelname~achieves similar performance on Common Set and Novel Instance Set for all three test settings. It shows that novel instances cause little impact on navigation and captioning ability. We attribute this instance generalization ability to the diverse training data, where most instances appear only one time, as shown in \cref{fig:freq_and_instance_num} (a). As for Novel Category Set, there is a small performance drop for navigation but an obvious performance drop for captioning. The navigator can be confused by an instance of novel categories, but may still find good viewpoints by observing other instances. However, for description generation, the captioner struggles to correctly describe instances whose categories have never been seen during training.   

\noindent\textbf{Potential benefits from joint navigation and captioning.}
Although our proposed model separates navigation and caption generation, we believe that simultaneously performing navigation and captioning would lead to better performance in a more efficient manner. 
The captioner can inform the navigator of objects, attributes, or relations that it is uncertain of, allowing the navigator to  explore the environment more efficiently in order to resolve the uncertainty and improve the quality of visual captioning.
We conduct an exploratory experiment to demonstrate the potential benefits of joint navigation and captioning.
We utilize the CaBOT captioner to generate scene descriptions at each navigation step predicted by the CaBOT navigation, and select the partial trajectory with the best CIDEr performance against the ground-truth captions.
As shown in~\cref{tab:joint_experiments}, this method achieves a significant improvement of +25.55 on the CIDEr and +28.02 on the CIDEr$^{l}$.
It indicates that imperfect navigation could bring more noise and deteriorate visual captioning, and thus it is necessary to guide the navigation according to the captioning ability.

\begin{table}
\caption{Breakdown analysis on different test subsets.}
    \label{tab:subset_experiments}
    \footnotesize
    \centering
    \tabcolsep=0.14cm
    \begin{tabular}{l|ccc|cc|cc}
    \toprule
    \multirow{2}*{\textbf{Subset}} & \multicolumn{3}{c|}{\textbf{Navigation}} & \multicolumn{2}{c|}{\textbf{OracleCap}} &  \multicolumn{2}{c}{\textbf{EmboCap}} \\
    ~ & NE$^{l}$$\downarrow$ & IS$^{l}$$\uparrow$ & SS$^{l}$$\uparrow$ & C$^{l}$$\uparrow$ &S$^{l}$$\uparrow$ & C$^{l}$$\uparrow$ &  S$^{l}$$\uparrow$ \\
    \midrule
    Common & \textbf{4.73} & 58.76 & 58.17 & 48.91 & 21.68 & 37.60 & 17.60 \\
    Novel Instance & 4.88 & \textbf{58.94} & \textbf{58.90} & \textbf{51.08} & \textbf{22.72} &  \textbf{39.10} & \textbf{18.45} \\
    Novel Category & 5.11 & 57.73 & 57.08 & 34.02 & 15.91 & 26.63 & 12.98 \\ 
    \bottomrule
    \end{tabular}
    \vspace{-1em}
\end{table}

\subsection{Qualitative Analysis}
\cref{fig:case_study} presents some embodied captioning results of our \modelname~on the test set. Case (a) shows that \modelname~can gradually approach instances and try to wrap around to find better viewpoints. When starting from a relatively lower position, \modelname~can also explore to find a higher viewpoint, as shown in case (b). \modelname~stops at a good viewpoint in case (b) but a bad viewpoint in case (a). However, it generates acceptable descriptions in both cases, which indicates that the overall quality of the trajectory matters more than the end viewpoints. 
We also show a failure case in case (c). When the agent reaches a bad viewpoint where no instance can be seen, it is less robust to find good viewpoints though historical visual and action information are leveraged. We consider this is mainly due to unseen states in testing, as we only utilize ground-truth trajectories for imitation learning. This problem can be relieved by data augmentation or reinforcement learning, and we leave it to future work.

\begin{table}
\caption{Exploratory experiments of joint navigation and captioning, where we simultaneously generate captions at each navigation step and stop navigation according to the captioning performance.}
    \label{tab:joint_experiments}
    \footnotesize
    \centering
    \begin{tabular}{p{0.2\linewidth}|p{0.07\linewidth}p{0.07\linewidth}p{0.07\linewidth}p{0.07\linewidth}}
    \toprule
    Method & CIDEr &  SPICE &  CIDEr$^{l}$ & SPICE$^{l}$ \\
    \midrule
    Cascade & 40.83 & 19.43 & 34.88 & 16.62  \\
    Joint & 66.38 & 21.51 & 62.90& 20.41 \\ 
    \bottomrule
    \end{tabular}
    \vspace{-1.5em}
\end{table}

\section{Conclusion}
In this work, we present a novel and challenging task called Embodied Captioning, where an agent should navigate in a 3D environment to gather visual information and  use natural language to comprehensively describe objects in the scene.
To support this task, we build the \dataset~dataset with manually annotated good viewpoints and paragraph descriptions for 10,000 synthetic scenes. 
We propose a Cascaded Embodied Captioning (CaBOT) model, which utilizes both visual and action history to perform navigation and then generates captions by leveraging the whole navigation trajectory. Experiments demonstrate the effectiveness of CaBOT and show a promising direction of joint modeling navigation and captioning.

There is no ethical or privacy issue with the \dataset~dataset as it uses publicly released simulated assets. As a result, the dataset is also limited in the simple simulated scenes.
The proposed Embodied Captioning task has the potential to benefit a wide range of applications, such as assisting visually impaired people.

{
\small
\noindent\textbf{Acknowledgements.}
This work was partially supported by the National Key  R\&D  Program  of  China  (No.2020AAA0108600) and the National Natural Science Foundation of China (No. 62072462). 
}

\appendix

We present more details about dataset construction and statistic in \cref{dataset_detail}. Details about the in-domain object detector, Template-based Captioner and implementation of CaBOT can be found in \cref{method_details}. The captioning ablation with predicted trajectory is presented in \cref{sec:cap_abla_pred_nav}. The detailed ablation study about how to leverage instance recognition knowledge from the object detector is shown in \cref{ablation_detail}. More qualitative results generated by CaBOT can be found in \cref{more_cases}. 

\section{Dataset Details}
\label{dataset_detail}

\subsection{Scene Simulation}
The details of the three steps for scene simulation are introduced below.

\noindent\textbf{Instance Resizing.} The size of 3D models loaded directly from Kurbic is not consistent with human common sense. For example, a `cup' instance can be as big as a `table' instance. To ensure scenes are more realistic, we design heuristic resizing rules for different categories of 3D models. For example, `bed' instances are resized to guarantee that the length is about 2.0m; `skateboard' instances are resized to a length close to 0.6m; the length of `bookshelf' instances is about 1.0m and their height should be lower than 2.0m. 

\noindent\textbf{Instance Selection.} 
Our indoor environment is organized by arranging small objects around big furniture such as beds, tables, etc. To simulate such indoor environment, we set one big furniture namely base instance, and multiple relatively smaller objects namely placing instances. We manually divide the categories of instances into base categories and placing categories.
For each scene, a base category and multiple placing categories are firstly selected. Due to the long tail distribution of different categories, simply sampling categories according to the number of instances under the category can lead to scenes with low diversity. Therefore, we empirically set a upper bound for sampling probability of each category when performing the category selection. 
Once the categories of a scene are determined, we randomly select an instance under each category.

\noindent\textbf{Instance Placing.} In addition to the category of instances, the position of instances is another crucial factor for scene diversity. 
It is difficult and tedious to manually design rules to place each instance. In this work, we perform physic simulation with the open-source PyBullet physics engine to place instances automatically. Concretely, the base instance is first placed in the center of the scene. The rest instances are then placed at 0.5m higher above the base instance and fall simultaneously due to gravity. Because of the collision between instances, the final orientations and locations of instances vary a lot across scenes.

\begin{figure}
    \centering
    \includegraphics[width=0.9\linewidth]{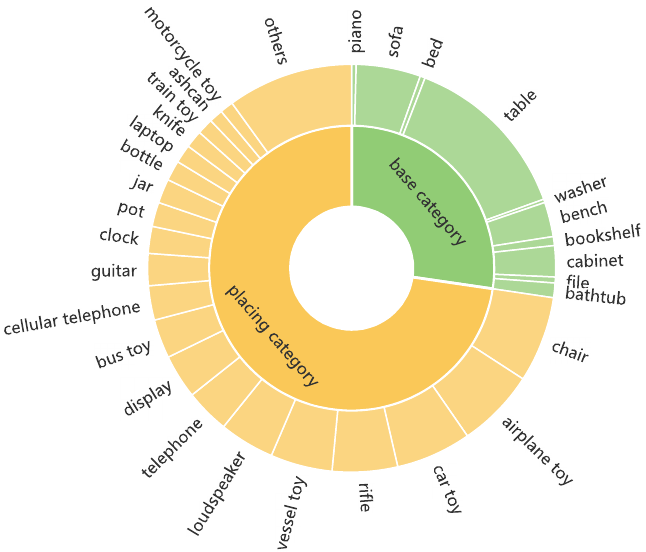}
    \caption{Category distribution of instances. 
    }
    \label{fig:categ_dis}
\end{figure}

\subsection{Annotation Details}
\begin{figure*}
    \centering
    \includegraphics[width=1.0\linewidth]{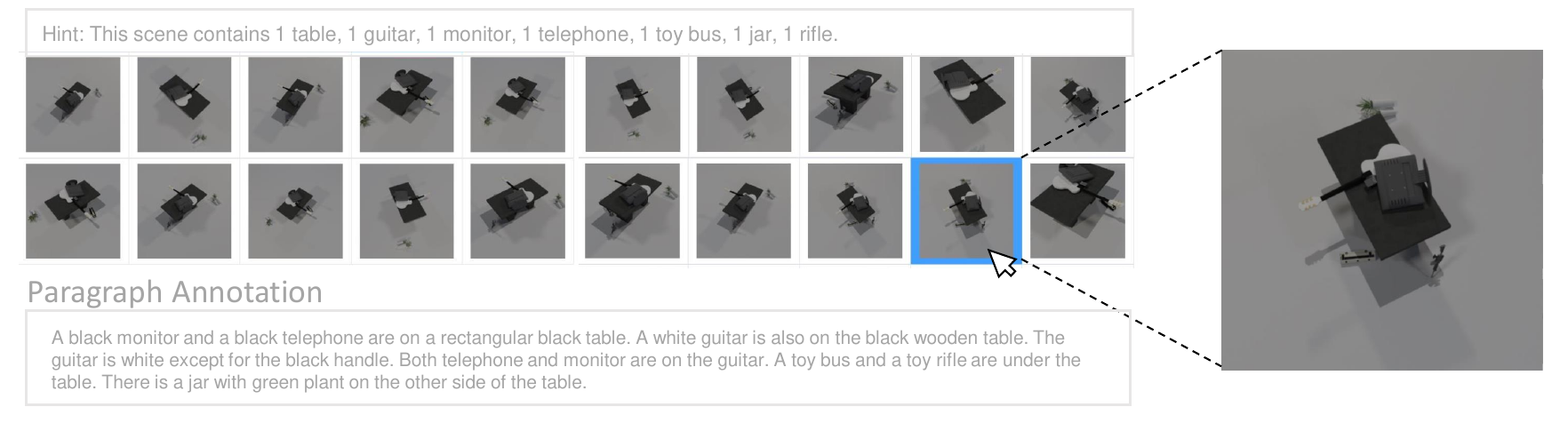}
    \caption{The annotation interface of viewpoint selection and paragraph annotation. The blue box denotes the selected viewpoint by the annotator. Each image can be enlarged to see instance details.}
    \label{fig:anno_platform}
\end{figure*}

\begin{figure*}
    \centering
    \includegraphics[width=1.0\linewidth]{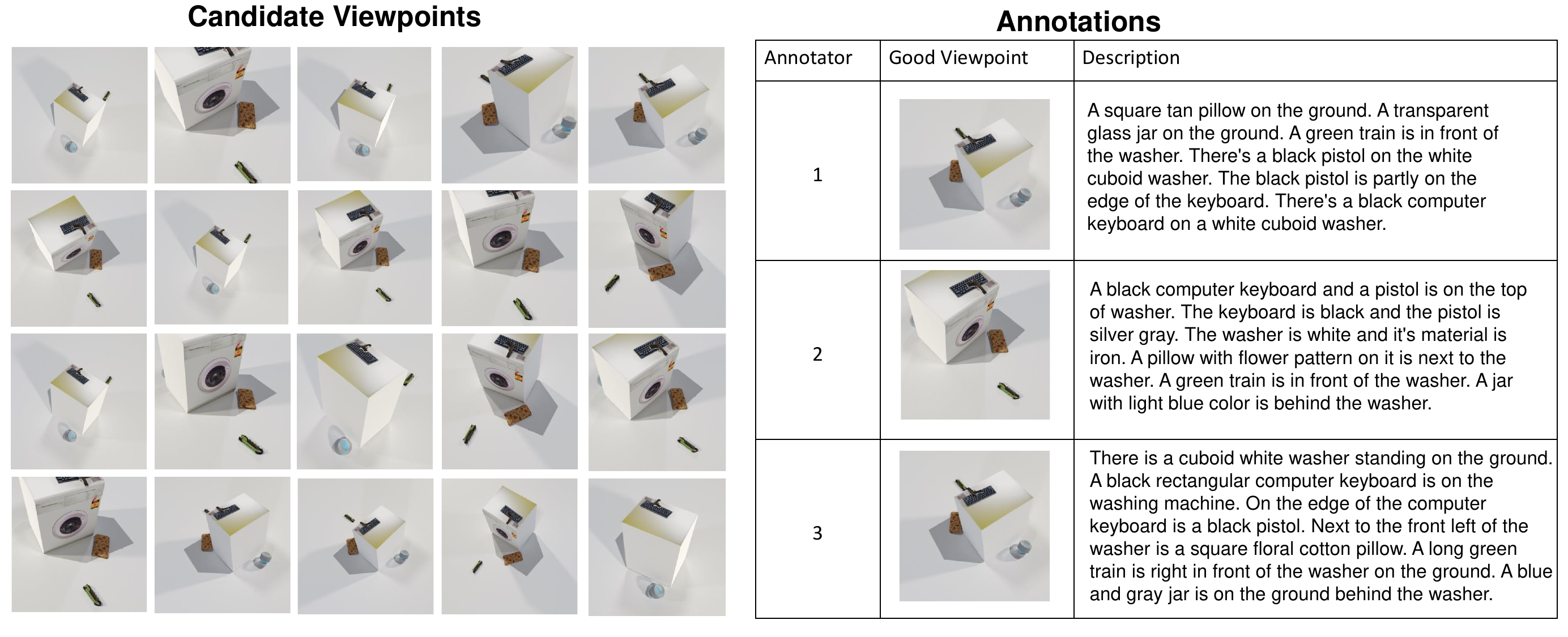}
    \caption{An example of good viewpoint selection and paragraph annotation by three annotators.}
    \label{fig:anno_sample}
\end{figure*}

The ET-Cap dataset contains good viewpoints and informative descriptions for each scene. 
\cref{fig:anno_platform} presents the annotation interface. For each scene, 
the annotator is provided with 20 images rendered from candidate viewpoints and a hint about the category of instances in the scene. The hint is a template \texttt{This scene contains [number$_1$] [category$_1$], [number$_2$] [category$_2$], ..., [number$_k$] [category$_k$].}" filled with instance categories and counts.
After selecting a good image, annotators should write a paragraph description at the bottom of the annotation interface. 
As shown in \cref{fig:anno_sample}, each scene is annotated by three workers. They may choose the same viewpoints but write different descriptions. 

In total, we hired 21 adults (8 males and 13 females) from 9 different cities to do the annotation. The annotators are proficient in English. It takes 2 months and costs about 5,700 dollars to complete annotating ET-Cap.

\begin{figure*}
    \centering
    \includegraphics[width=0.9\linewidth]{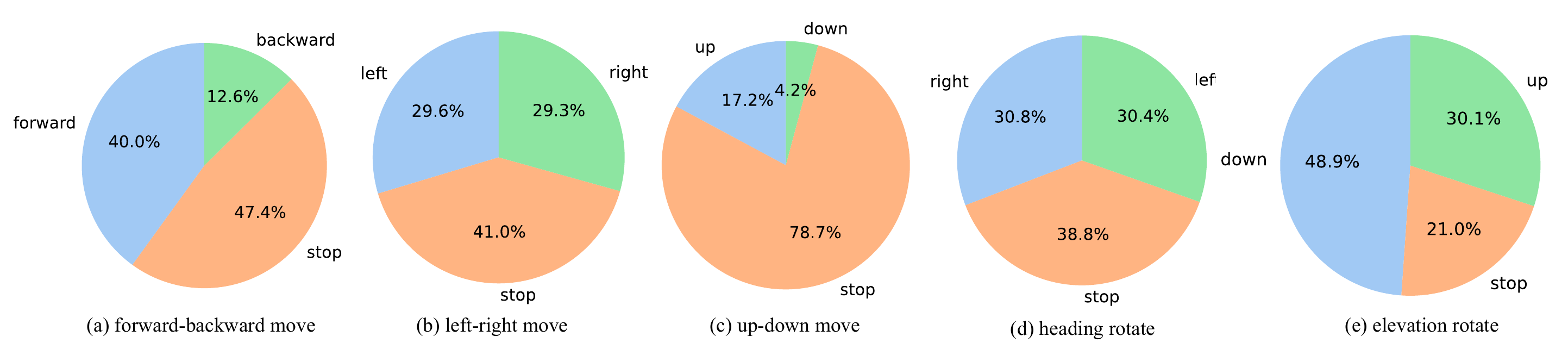}
    \caption{Distribution of forward-backward move actions (a), left-right move actions (b), up-down move actions (c), heading rotate actions (d) and elevation rotate actions (e).
    }
    \label{fig:action_dis}
\end{figure*}

\subsection{Dataset Statistics}
As shown in Fig.~\ref{fig:categ_dis}, instance categories are diverse overall. The `table' instance is relatively more than other categories of instances because it is the most common indoor furniture. Fig.~\ref{fig:action_dis} shows the direction distribution of each action type in ground-truth trajectories.

\section{Method Details}
\label{method_details}

\begin{table*}
\caption{Captioning ablation study with trajectories predicted by the CaBOT navigator.
`single' means only utilizing the end view for region-level cross-attention, and `multi' means using the mean view of all observations. }
    \label{tab:cap_abla_pred_nav_experiments}
    \footnotesize
    \tabcolsep=0.08cm
    \centering
    \begin{tabular}{c|ccc|ccccc|ccccc}
    \toprule
    ~  & Region-level & Trajectory-Level & Init & \multirow{2}*{BLEU4} & \multirow{2}*{METEOR} & \multirow{2}*{ROUGE-L} & \multirow{2}*{CIDEr}  & \multirow{2}*{SPICE} & \multirow{2}*{BLEU4$^{l}$} & \multirow{2}*{METEOR$^{l}$} & \multirow{2}*{ROUGE-L$^{l}$} & \multirow{2}*{CIDEr$^{l}$}  & \multirow{2}*{SPICE$^{l}$} \\
     ~  & CrossAtt & CrossAtt & Backbone & ~ & ~ & ~ & ~ & ~ & ~ &   \\
    \midrule
    r1 & single &$\times$ & $\times$ & 24.71 & 20.31	& 44.37	& 27.14	&  13.66 & 	18.58& 	18.20 & 	37.96 & 23.22 &  11.70 \\
    r2 & multi & $\times$ & $\times$  & 24.70 & 22.06& 43.38&	28.50 &	16.06	&16.87	&19.37	&37.10	&24.34	&13.74 \\
    r3  & $\times$ & $\checkmark$ & $\times$  & 25.40	& 23.16 & 	44.35 &	36.98	& 17.60 &	18.37 &	20.47 &	37.92 &	31.49 &	15.05  \\
    r4 & single & $\checkmark$ & $\times$ & 24.97 &	22.18 &	43.66 &	29.00 &	16.47 &	17.49 &	19.50 &	37.39 & 24.80 &	14.12 \\ 
    r5 & multi & $\checkmark$ & $\times$ &  26.13 &	23.14 &	45.09 &	37.87 &	18.36 &	18.98 &	20.44 &	38.57 &	32.29 &	15.72 \\
    r6 & multi & $\checkmark$ & $\checkmark$ & \textbf{26.45} & \textbf{23.56} & \textbf{45.48} & \textbf{40.83} & \textbf{19.43} & \textbf{19.22} & \textbf{20.82} & \textbf{38.87} & \textbf{34.88} & \textbf{16.62} \\
    \bottomrule
    
    \end{tabular}
\end{table*}

\subsection{In-domain Object Detector Details}
To study whether the object detection ability contributes to Embodied Captioning task, we train an in-domain DETR\cite{Carion2020detr} model and leverage its parameters in our model, as mentioned in Sec.~4.1 and Sec.~4.2. To train the in-domain detector,  we construct extra 4,398 scenes like ET-Cap and render 87,960 images with object category and bounding boxes annotation given by Kubric simulator. The model is trained for 46 epochs and achieves 39.4 mAP on a test set, which includes 150 scenes with 3,000 images.

\subsection{Baseline Details}

\noindent\textbf{Template-based Captioner} first generates descriptions for each viewpoint by fill-in templates and then merges them into a trajectory description. For each viewpoint, it detects salient objects with the in-domain DETR model. Objects with a confidence score higher than 0.9 are selected to construct the description. We attach each object with a color attribute according to its RGB values. After detecting the objects, it first describes the largest object \texttt{[obj$_L$]} by selecting a template like "\texttt{There is a [obj$_{L}$] in the scene.}". Then it describes other objects one by one in descending order of bounding box size. For a smaller object \texttt{[obj$_{s}$]}, it describes its spatial relation with a randomly selected larger object \texttt{[obj$_{l}$]} with templates such as \texttt{A/An [obj$_{s}$] is [Relation] the [obj$_{l}$].}". Note that \texttt{[obj$_{l}$]} has occurred in one of the former sentences. The spatial relation is predicted by the relative positions between objects similar to Preposition Functions~\cite{babytalk}. After generating captions for each viewpoints, it further ensembles all captions into a trajectory description as follows: 1) choose the best viewpoint with the highest summation of object confidences, and treat its description as the basic description; 2) describe objects that do not occur in the best viewpoint with a template "\texttt{There is also a [obj$_1$], a [obj$_2$], ..., and a [obj$_n$] in the view.}". We design multiple synonymous templates according to the style of the ground-truth paragraphs as shown in \cref{tab:templates}. We randomly select one template for each object during inference. 
\begin{table}[h]
\caption{Templates used in template-based captioner.}
\centering
\scalebox{0.75}{
\begin{tabular}{l|l}
\toprule
Describe objects & \multicolumn{1}{c}{Templates} \\ \midrule
\multirow{7}{*}{Largest} & \texttt{A/An [obj$_L$] in on the ground.} \\
 & \texttt{A/An [obj$_L$] is in the view.} \\
 & \texttt{There is a/an [obj$_L$].} \\
 & \texttt{There is a/an [obj$_L$] on the ground.} \\
 & \texttt{There is a/an [obj$_L$] in the view.} \\
 & \texttt{On the ground is a/an [obj$_L$].} \\
 & \texttt{In the view there is a/an [obj$_L$].} \\ \midrule
\multirow{2}{*}{Others} & \texttt{A/An [obj$_{s}$] is [Relation] the [obj$_{l}$].} \\
 & \texttt{[Relation] the [obj$_{l}$] is a/an [obj$_{s}$].} \\ \midrule
 \multirow{2}{*}{Not occurs} & \texttt{There is also a [obj$_1$], a [obj$_2$], ...,} \\
 & \texttt{and a [obj$_n$] in the view.} \\ \bottomrule
\end{tabular}
}
\label{tab:templates}
\end{table}

\subsection{Implementation Details} 
In the navigator, the number of transformer layers of the Region Encoder, Historical Vision Encoder and Historical Action-informed Decoder is 2, 1, 1, respectively. The transformer layer number of the Bi-CrossAttention Decoder in the captioner is set as 2. The maximum trajectory length is set as 12. The maximum caption length is set as 77. The navigation module and the captioning module are trained with 10 epochs and 20 epochs, respectively. Both two modules are trained with the AdamW \cite{Loshchilov2019adamw} optimizer. The learning rate for the navigation module and captioning module is $1e^{-4}$ and $3e^{-5}$, respectively. The batch size of these two modules is set as 32.

\begin{figure*}
    \centering
    \includegraphics[width=1.0\linewidth]{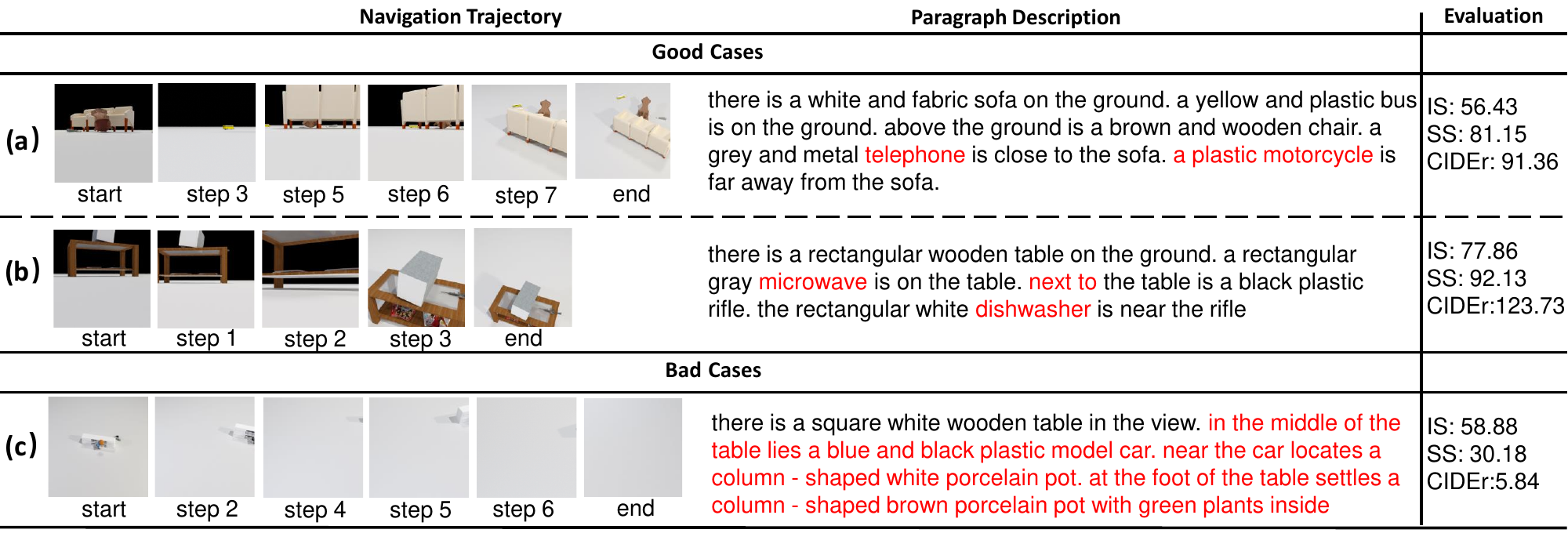}
    \caption{More qualitative results of CaBOT on Embodied Captioning test set. Wrong descriptions are marked in {\color{red}red}.}
    \label{fig:case_study_supp}
\end{figure*}

\section{Captioning Ablation with Predicted Trajectory}
\label{sec:cap_abla_pred_nav}
In Sec 5.2, we perform captioning ablation study with oracle trajectory to verify the effectiveness of our Bi-CrossAttention Decoder and DETR initialization. To verify that they are also crucial for Embodied Captioning, we further perform an ablation study with trajectory predicted by the CaBOT navigator, as shown in \cref{tab:cap_abla_pred_nav_experiments}. Firstly, utilizing the mean view of the predicted trajectory outperforms the one utilizing only the end view (r2 vs r1), especially on CIDEr and SPICE. Besides, leveraging the predicted trajectory at the trajectory level is better than at the region level (r3 vs r2). These also support that merging visual information at the trajectory level is better for scene description.  Besides, Bi-CrossAttention outperforms single region-level or trajectory-level cross-attention (r5 vs r2, r5 vs r3), initializing the backbone with pre-trained DETR improves the captioning performance (r6 vs r5), both of which are also consistent with the observations in Sec 5.2.

\begin{table}
\caption{Ablation study about initialization and training strategy. Oracle-trajectory Captioning are tested withthe captioner of CaBOT. `DeInit' means whether using transformer encoder of DETR to initialize region encoders of the navigator or the captioner.`RN' and `EN' refer to ResNet and Transformer Encoder of DETR.}
    \label{tab:detr_init_experiments}
    \footnotesize
    \centering
    \begin{tabular}{p{0.01\linewidth}|p{0.17\linewidth}p{0.08\linewidth}|p{0.05\linewidth}p{0.08\linewidth}|p{0.09\linewidth}p{0.09\linewidth}}
    \toprule
     &\multicolumn{2}{c|}{\textbf{Backbone}} & \multicolumn{2}{c|}{\textbf{Region En}}  & \multicolumn{2}{c}{\textbf{OracleCap}} \\
    & Init & Frozen & DeInit & Frozen  & BLEU4$\uparrow$ & CIDEr$\uparrow$ \\
    \midrule
    r1 & ResNet50 & $\times$  & $\times$ & $\times$  & 27.58 & 42.10 \\
    r2 & Detr RN & $\checkmark$  & $\times$ & $\times$  & 27.30 & 40.91 \\
    r3 & Detr RN & $\times$  & $\times$ & $\times$ & 27.62 & \textbf{45.24} \\
    r4 & Detr RN & $\times$  & $\checkmark$ & $\times$ & \textbf{27.94} & 44.85 \\
    r5 & Detr RN & $\checkmark$  & $\checkmark$ & $\checkmark$ & 26.33 & 35.30 \\
    r6 & Detr RN+En & $\times$  & $\times$ & $\times$ & 27.08 & 43.73 \\
    r7 & Detr RN+En & $\checkmark$  & $\times$ & $\times$ & 26.78 & 38.66 \\
    \bottomrule
    \end{tabular}
\end{table}

\section{Ablation of Instance Recognition Ability}
\label{ablation_detail}

As mentioned in Sec.5.2, knowledge about object recognition from the in-domain DETR model boosts captioning performance. In this section, we compare different strategies of initialization and training to explore how to best leverage this knowledge for Embodied Captioning. As show in Table \ref{tab:detr_init_experiments}, for the captioner, initializing its backbone with only ResNet of DETR and optimizing it during training is best (r3). It indicates that for captioning, besides category and attributes of each instance, it is also crucial to understand their spatial relationships. Therefore, only leveraging basic knowledge of object detection model and optimizing it during captioning training benefits the captioner more.

\section{More Qualitative Results}
\label{more_cases}
\cref{fig:case_study_supp} shows more qualitative results given by CaBOT. The case (a) shows that the navigator could explore the environment and find a small and relatively far away instance (the yellow toy bus). The captioner could combine vision information  provided from different viewpoints to generate the description. In case (c), the agent reaches bad viewpoints during navigation and can not find better viewpoints or return to original viewpoints anymore. This shows that though historical vision and camera information have been leveraged by the navigator, it may still be confused when no instance is visible. Besides, when there are little visual instance information provided in the trajectory, the captioner randomly generates some irrelevant descriptions.

{\small
\bibliographystyle{ieee_fullname}
\bibliography{reference}
}

\end{document}